\documentclass[letterpaper, 10 pt, conference]{ieeeconf}  

\IEEEoverridecommandlockouts                              

\usepackage{amsmath}
\usepackage{graphicx}
\usepackage{amssymb}
\usepackage{booktabs}
\usepackage{multirow}
\usepackage{multicol}
\usepackage{float}
\usepackage{subfigure}
\usepackage{kotex}
\usepackage{cite}
\usepackage{tabularx}
\usepackage{array}
\usepackage[flushleft]{threeparttable}
\usepackage[ruled,vlined]{algorithm2e}
\usepackage[dvipsnames]{xcolor}
\usepackage{dblfloatfix} 

\pdfoutput=1

\begin{document}
%
\title{\bf WALK-VIO: Walking-motion-Adaptive Leg Kinematic Constraint \\ Visual-Inertial Odometry for Quadruped Robots}
%
%
%

\author{Hyunjun Lim$^{1}$, Byeongho Yu$^{1}$, Yeeun Kim$^{1}$, Joowoong Byun$^{2}$, Soonpyo Kwon$^{2}$, \\ Haewon Park$^{2}$, and Hyun Myung$^{*}$, \IEEEmembership{Senior~Member, IEEE}%
\thanks{$^{1}$H. Lim, B. Yu, and Y. Kim are with School of Electrical Engineering, Korea Advanced Institute of Science and Technology~(KAIST), Daejeon, Republic of Korea
{\tt\small \{tp02134, bhyu, yeeunk\}@kaist.ac.kr}}%
\thanks{$^{2}$J. Byun, S. Kwon, and H. Park are with Department of Mechanical Engineering, KAIST, Daejeon, Republic of Korea
{\tt\small \{woong164, happymen80, haewonpark\}@kaist.ac.kr}}
\thanks{$^{*}$H. Myung is with School of Electrical Engineering, KI-AI, and KI-R, KAIST, Daejeon, Republic of Korea
{\tt\small hmyung@kaist.ac.kr}}%
\thanks{This work was supported by the Defense Challengeable Future Technology Program of Agency for Defense Development, Republic of Korea. The students are supported by Korea Ministry of Land, Infrastructure and Transport (MOLIT) as “Innovative Talent Education Program for Smart City” and BK21 FOUR.}
}

%
%

\markboth{IEEE Robotics and Automation Letters. Preprint Version. Accepted Month, Year}
{FirstAuthorSurname \MakeLowercase{\textit{et al.}}: ShortTitle} 
%



\maketitle

\begin{abstract}
In this paper, WALK-VIO, a novel visual-inertial odometry (VIO) with walking-motion-adaptive leg kinematic constraints that change with body motion for localization of quadruped robots, is proposed. Quadruped robots primarily use VIO because they require fast localization for control and path planning. However, since quadruped robots are mainly used outdoors, extraneous features extracted from the sky or ground cause tracking failures. In addition, the quadruped robots' walking motion cause wobbling, which lowers the localization accuracy due to the camera and inertial measurement unit (IMU). To overcome these limitations, many researchers use VIO with leg kinematic constraints. However, since the quadruped robot's walking motion varies according to the controller, gait, quadruped robots' velocity, and so on, these factors should be considered in the process of adding leg kinematic constraints. We propose VIO that can be used regardless of walking motion by adjusting the leg kinematic constraint factor. In order to evaluate WALK-VIO, we create and publish datasets of quadruped robots that move with various types of walking motion in a simulation environment. In addition, we verified the validity of WALK-VIO through comparison with current state-of-the-art algorithms.
\end{abstract}


%
\IEEEpeerreviewmaketitle
\section{Introduction}

%
%
%
%

\begin{figure}[t]
    \centering
    \subfigure[]{\label{fig:feature_champ}\includegraphics[width=0.49\linewidth]{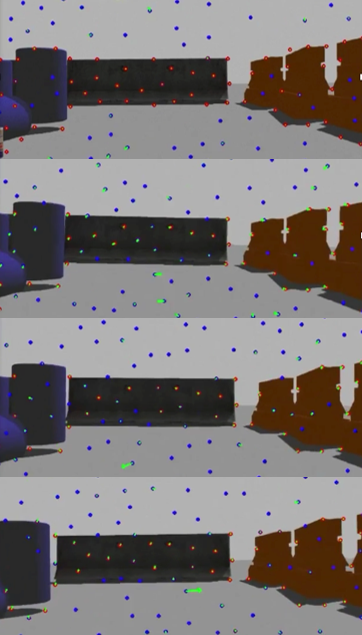}}
    \subfigure[]{\label{fig:feature_NMPC}\includegraphics[width=0.49\linewidth]{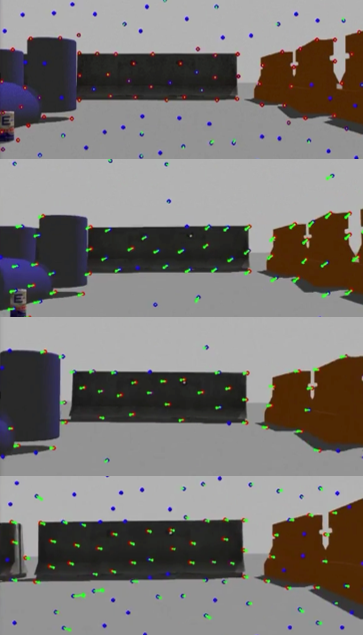}}
    \caption{Images of extracted point features' movement on ANYmal \cite{hutter2016anymal} using \subref{fig:feature_champ} Champ \cite{champ} and \subref{fig:feature_NMPC} NMPC (nonlinear model predictive control) \cite{hong2020real}. The green lines are the movement of features from the previous frame to the current frame. The walking motion of the quadruped robot varies depending on the controller, which can be determined by the movement of features. When moving with a large walking motion, the moving distance of point features increases.}
    \label{fig:feature}
\end{figure}

Through\textcolor{black}{out} the development of the robot industry, \textcolor{black}{research} on mobile robots \textcolor{black}{has} been actively conducted, and in recent years, many quadruped robots have been developed. Wheeled robots have the disadvantage that they can only be driven in environments such as \textcolor{black}{on} roads, while quadruped robots can move stably without restrictions on the environment. However, quadruped robots require more complex control and path planning technology to move than wheeled robots do. In addition, wheel-type robots have wheel odometry, whereas quadruped robots require localization technology based on other sensors such as joint encoders and torque sensors.

\textcolor{black}{A} quadruped robot's localization system mainly uses the visual-inertial odometry (VIO), which combines a camera and an inertial measurement unit (IMU). Quadruped robots require fast localization because control and path planning are more complicated than \textcolor{black}{for} other mobile robots. Therefore, cameras are more suitable for quadruped robots than LiDARs, which use many point clouds. In addition, since four legs must support the weight, the robot's weight can be reduced by using a camera  which is light compared to other sensors.

Several difficulties arise \textcolor{black}{when} using VIO in a quadruped robot. First, quadruped walking is mainly \textcolor{black}{used} outdoors, \textcolor{black}{where the environment includes} textureless areas such as the sky and the \textcolor{black}{ground}. When features are extracted from the sky and the \textcolor{black}{ground}, localization accuracy is degraded due to the point feature's tracking failure. In addition, the camera image \textcolor{black}{is blurred due to the motion} of the robot\textcolor{black}{'s walk}. While general unmanned ground vehicle (UGV) and unmanned aerial vehicle (UAV) movements have little wobbling, quadruped robots have repetitive wobbling due to the walking process. 

To overcome the difficulties mentioned above, \textcolor{black}{researchers have studied} using VIO with leg kinematics. Leg kinematics uses joint encoder data to calculate information such as the position and velocity of each leg's end effector. If the end effector of the leg touches the ground, the body's movement can be calculated by fixing the end effector. This can be used as a new constraint to complement the VIO of the quadruped robot.

Unlike other mobile robots, quadruped robot\textcolor{black}{s'} VIO is affected by walking motions, which varies with several factors such as controller, gait, speed, and so on. As shown in Fig.~\ref{fig:feature}, the walking motion varies depending on the controller, and can be determined through the movement of features. In general, the large walking motion has the advantage of \textcolor{black}{allowing quick movement in} challenging environment\textcolor{black}{s}. However, the large walking motion \textcolor{black}{creates} vibration and impact, making \textcolor{black}{the use of} camera and IMU measurement relatively inaccurate. Conversely, since leg kinematics is not affected by walking motions, it can be used as an important measurement even with a large walking motion.

However, since the existing VIO algorithms applying leg kinematics do not consider various walking motions, they cannot guarantee robustness. In this paper, we propose walking-motion-adaptive leg kinematic constraints-based VIO, called WALK-VIO, for quadruped robots. The main contributions of this study are as follows:
\begin{itemize}
    \item For the localization of the quadruped robots, we implement robust optimization-based VIO with leg kinematic constraints. 
    \item \textcolor{black}{We propose a} novel adaptive factor to compensate the walking motion in quadruped robots' localization.
    \item We create and publish simulation datasets \textcolor{black}{that consider} the actual quadruped robot model and various walking motions for localization.
\end{itemize}

The remainder of this paper is organized as follows. Section~\ref{sec:related work} provides an overview of related works. Section~\ref{sec:proposed method} describes the proposed \textcolor{black}{WALK-VIO} in detail. Section \ref{sec:experimental results} presents the experimental results. Finally, Section \ref{sec:conclusion} summarizes our contributions and discusses future work.

\section{Related works}\label{sec:related work}
\textcolor{black}{Researchers have considered} the use of various sensors to estimate the state of the robot. No complete sensor configuration exists, and the user selects the \textcolor{black}{appropriate} sensor according to the required performance and operating environment. Although there are various combinations, the \textcolor{black}{types of} sensors covered in this paper are limited to the camera, which is a light\textcolor{black}{-}weight and low\textcolor{black}{-}cost sensor, joint encoders that measure the joint rotation speed, and an IMU.

Likewise, the \textcolor{black}{user chooses the} platform according to the situation. Mobile robots are preferred \textcolor{black}{for use} on flat \textcolor{black}{surfaces} such as indoors, and UAVs are often used when \textcolor{black}{a significant amount of movement} within a short time \textcolor{black}{is required}. In addition, quadruped robots are used in environments with ground that are not flat and \textcolor{black}{have} many obstacles, such as in a field or a forest. In such environment\textcolor{black}{s}, it is not easy to use measurements such as camera\textcolor{black}{s} and IMU for localization. Therefore, attempts have been made to improve the localization performance through additional constraints in quadruped robots.

For localization of quadruped robots, filtering methods have been widely used for a long time. Most studies performed sensor fusion using a Kalman filter. \textcolor{black}{In a representative }study, \textcolor{black}{researchers} carried out state estimation by mounting an extended Kalman filer (EKF)-based algorithm on \textcolor{black}{Boston Dynamics'} LS3 robot \cite{ma2016real}. \textcolor{black}{Predictions were} performed through inertial measurements and updated through visual measurements. In their study, leg odometry \textcolor{black}{was} also used, but only when visual odometry \textcolor{black}{had failed} to estimate. When state estimation is performed using a filter, only the most recent state is used, and the previous states are not preserved.

\textcolor{black}{In contrast}, the smoothing method \textcolor{black}{uses all past states and can be highly accurate} compared to the filtering method because global optimization is possible. \textcolor{black}{R}epresentative studies \textcolor{black}{include one that estimated} the relative motion through SVO\cite{forster2016svo} and \textcolor{black}{one that used} iSAM2 \cite{kaess2012isam2}, widely used back-end algorithms, for a biped robot \cite{hartley2018hybrid}. Because of the wobbling or shock that occurs when quadruped robot\textcolor{black}{s} move, the pre-integration\cite{forster2016manifold} method, which is mainly used in the UAV community, leads to inaccurate results. In order to improve the results, researchers used a square-root unscented Kalman filter (SR-UKF) in parallel with VINS in COCLO\cite{yang2019state}. Position and orientation estimates \textcolor{black}{were} obtained through the output of SR-UKF, and the output of VINS \textcolor{black}{was} used to update the SR-UKF state. \textcolor{black}{In a recent} paper, \textcolor{black}{researchers} combine\textcolor{black}{d} both \textcolor{black}{types of} inertial information, visual and leg odometry\textcolor{black}{,} through factor graph optimization\cite{wisth2019robust}. Experiments were conducted in an outdoor environment, but various gait or rapid movement verification was not performed.

Moreover, due to quadruped robots' characteristics, the controller or state estimation method is often dependent on the hardware configuration. Algorithms specialized in RHex\cite{saranli2001rhex} have been proposed, but \textcolor{black}{they are limited by} poor generality \cite{lin2005leg,lin2006sensor,skaff2010context}. \textcolor{black}{Researchers proposed a} method of expanding the \textcolor{black}{algorithm's} versatility to apply leg odometry to various gait\textcolor{black}{s} \cite{reinstein2011dead}. However, according to \cite{kubelka2012complementary}, \textcolor{black}{this method is limited by the large amount of computation required}. In addition, \textcolor{black}{another} study that conducted state estimation of StarlETH\cite{hutter2012starleth,bloesch2013state1,bloesch2013state2} \textcolor{black}{combined} joint encoder and IMU measurement using EKF to estimate the full state of \textcolor{black}{a} quadruped robot. However, as mentioned above, EKF \textcolor{black}{is less accurate than} the graph optimization method.

\begin{figure*}[t]
    \centering
    \includegraphics[width=1.0\linewidth]{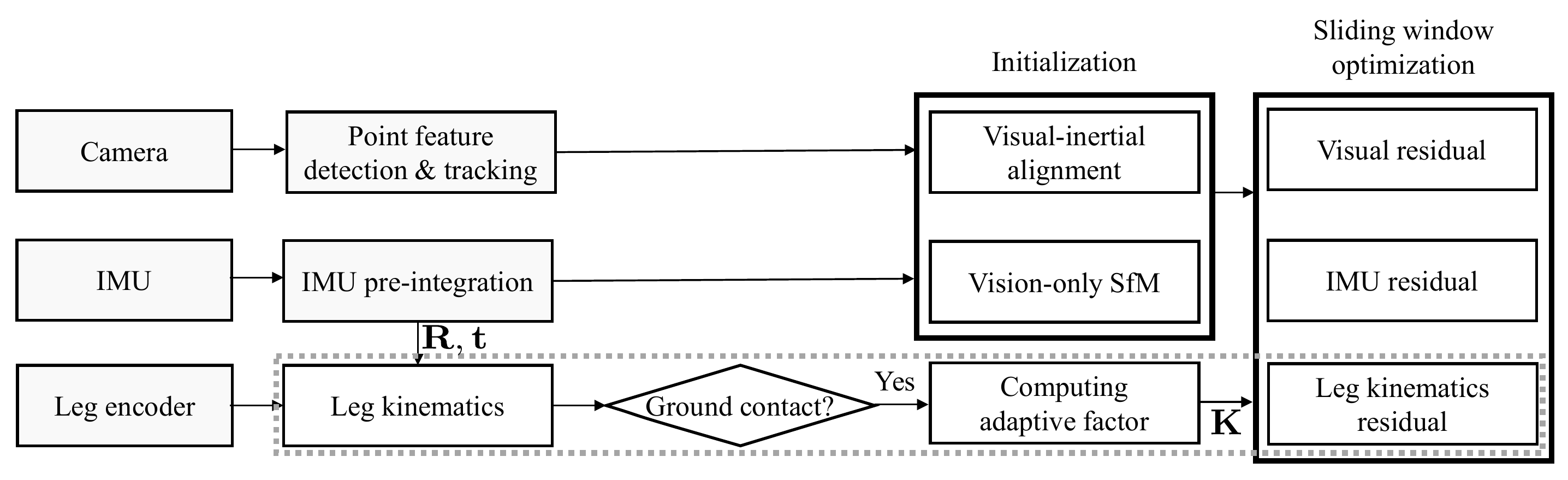}
    \caption{Block diagram illustrating the framework of \textcolor{black}{WALK-VIO}. The dashed box represents the newly added blocks in this paper. Leg kinematics is calculated based on the pre-integration result of the IMU and is explained in Section~\ref{sec:proposed method}.\textit{A}. If the leg is in \textcolor{black}{a} contact state, an adaptive factor is calculated using the motion of the feature, as explained in Section~\ref{sec:proposed method}.\textit{B}. Finally, the leg kinematic residual defined with the adaptive factor is added to VIO, as explained in Section~\ref{sec:proposed method}.\textit{C}.}
    \label{fig:framework}
\end{figure*}

\section{Proposed Method}\label{sec:proposed method}
\textcolor{black}{WALK-VIO} extends the VINS-Fusion\cite{qin2018vins, qin2019general} by adding leg kinematic constraints, as shown in Fig.~\ref{fig:framework}. The dashed box represents the newly added blocks in this paper. Because this algorithm uses VINS-Fusion as a baseline, IMU measurements use the pre-integration method \cite{forster2016manifold}; optimization uses two-way marginalization with \textcolor{black}{the} Schur complement \cite{sibley2010sliding}; and point feature manipulation uses optical-flow, triangulation, and re-projection error as with the VINS-Fusion.

\subsection{Leg Kinematic Constraint}\label{sec:leg kinematic}
\textcolor{black}{Leg} kinematics calculates the pose of each leg from the start point to the end effector. The pose from the start point of the leg to the end effector, $\mathbf{s}$, calculated through the angles of the encoders installed in the leg is as follows:
\begin{equation}
\mathbf{s} = \mathbf{lk}(\boldsymbol{\theta}) + \mathbf{n}_{\textrm{lk}},
\end{equation}
where $\mathbf{lk}$ denotes the leg kinematic model, and $\boldsymbol{\theta}$ is the joint angle vector. At this time, since the joint angle vector has Gaussian noise, the noise for leg kinematics is defined as $\mathbf{n}_{\textrm{lk}}$. In this paper, \textcolor{black}{we used} the Kinematics and Dynamics Library (KDL) \cite{bruyninckx2001open} in calculating leg kinematics.

\begin{figure}[t]
    \centering
    \includegraphics[width=\linewidth]{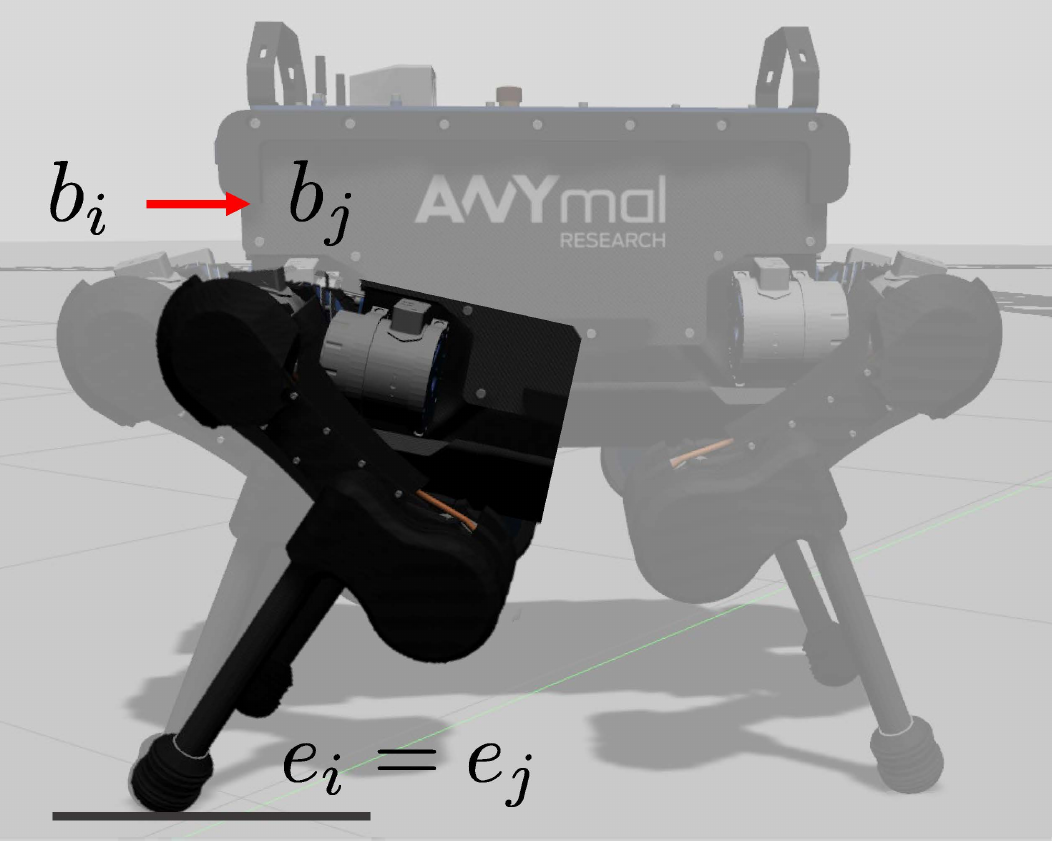}
    \caption{When \textcolor{black}{the robot moves} from the body state $b_i$ to the body state $b_j$, the position of the end effector ($e_i$ and $e_j$) does not change in contact with ground. Thus, it is possible to estimate the pose change of the body state through the change in the leg kinematics.}
    \label{fig:joint_measurement}
\end{figure}

The kinematic constraints we define can be used in ground contact conditions. Assuming the robot moves from the body state $b_i$ to the body state $b_j$, when a ground contact condition is detected in two adjacent frames, i.e., the $i$-th and the $j$-th frames in the sliding window used in VIO, the end effector position in the two frames does not change, as shown in Fig.~\ref{fig:joint_measurement}. Therefore, the difference from the IMU prediction through kinematic measurement is defined as a residual as follows:
\begin{equation}
\begin{split}
\mathbf{r}_{\mathcal{L}_{ij}} &= \mathbf{s}_j  - \mathbf{\tilde{t}}^{b_j}_{e_j} \\
&= \mathbf{s}_j - (\mathbf{\tilde{R}}^w_{b_j})^T(\mathbf{\tilde{t}}^w_{e_j} - \mathbf{\tilde{t}}^w_{b_j}) \\
&= \mathbf{s}_j - (\mathbf{\tilde{R}}^w_{b_j})^T(\mathbf{\tilde{t}}^w_{e_i} - \mathbf{\tilde{t}}^w_{b_j}) \\
&= \mathbf{s}_j - (\mathbf{\tilde{R}}^w_{b_j})^T(\mathbf{\tilde{R}}^w_{b_i}\mathbf{s}_i + \mathbf{\tilde{t}}^w_{b_i} - \mathbf{\tilde{t}}^w_{b_j}),
\end{split}
\end{equation}
where $\mathbf{\tilde{R}}^w_b$ and $\mathbf{\tilde{t}}^w_b$ are estimations of quaternion and position of body state based on world coordinates from IMU's pre-integration, respectively. In addition, $\mathbf{\tilde{t}}^b_e$ and $\mathbf{\tilde{t}}^w_e$ are translation estimations from body and world coordinates to the end effector coordinate, respectively.

\subsection{Adaptive Factor Analysis}
When a quadruped robot has a large walking motion, the camera and IMU measurements are affected, while the joint encoder is not. When the walking motion is large, the moving distance of the continuously observed features is large. Therefore, it degrades the performance of KLT tracking \cite{lucas1981iterative, shi1994good}, which is based on the optical flow mainly used in VIO. In addition, as explained in \cite{yang2019state}, the IMU's pre-integration \textcolor{black}{yieids} inaccurate results due to the walking movement. Conversely, the joint encoder can be considered a relatively more important constraint when the camera and IMU are unstable, because it has the similar noise level regardless of the walking motion.

In order to change the importance of leg kinematic constraints according to the walking motion of the quadruped robot, we propose a new factor in the graph optimization process. This new factor uses the feature's motion to consider the vibration and wobbling of the quadruped robot. Because the vibration and wobbling of a quadruped robot is mainly swaying left and right, it is simpler to predict the motion based on the movement distance of the features than to do so using IMU or leg kinematics. In order to determine the walking motion of the quadruped robot, we calculate the average motion of the point features. 

The average movement of features, $\mathbf{d}_i$, observed in the $i$-th frame of the sliding window is as follows:
\begin{equation}
\begin{aligned}
\mathbf{d}_i 
&= (d_{i,x}, d_{i,y}) \\
&= ({\sum_{k=1}^{n}(u^k_i-u^k_{i-1})\over{n}}, {\sum_{k=1}^{n}(v^k_i-v^k_{i-1})\over{n}}),
\end{aligned}
\end{equation}
where $n$ is the number of point features; $u^k_i$ and $v^k_i$ are the image's $x$ and $y$ coordinates of the $k$-th feature in the $i$-th frame of the sliding window, respectively.

Principal Component Analysis (PCA)\cite{wold1987principal} is used to analyze the motion of 2D features. PCA can calculate the principal component vector by calculating the covariance matrix of the data. Considering the extrinsic parameters in this process, the camera image's $x$ and $y$ coordinates are the same as the $y$ and $z$ coordinates of the body state. Therefore, the covariance of average movement of features, $\mathbf{C}$, observed in the $i$-th frame of the sliding window is obtained as follows:
\begin{equation}
\begin{gathered}
\mathbf{C} =
\begin{bmatrix}  
1 & 0 & 0 \\
0 & c_{xx} & c_{xy} \\
0 & c_{xy} & c_{yy}
\end{bmatrix}, \\
c_{xx} = \sum_{i=1}^{N}(d_{i,x} - \Bar{d}_x)^2, \\ 
c_{xy} = \sum_{i=1}^{N}(d_{i,x} - \Bar{d}_x)(d_{i,y} - \Bar{d}_y), \\
c_{yy} = \sum_{i=1}^{N}(d_{i,y} - \Bar{d}_y)^2,
\end{gathered}
\end{equation}
where $N$ is the size of the sliding window; $\Bar{d}_x$ and $\Bar{d}_y$ are means of average moving distances in the image's $x$- and $y$-axes, respectively. Through decomposition of the covariance, the leg kinematic residual factor, $\mathbf{\Gamma}$, is obtained as follows:
\begin{equation}
\begin{gathered}
\mathbf{C} = \mathbf{P}{\mathbf{\Lambda}}\mathbf{P}^T, \\
\mathbf{\Gamma} = \mathbf{P}\mathbf{\Lambda}^{1/2}, \\
\mathbf{\Lambda} = \begin{bmatrix}
\lambda_x & 0 \\
0 & \lambda_y
\end{bmatrix},
\end{gathered}
\end{equation}
where $\mathbf{P}$ and $\mathbf{\Lambda}$ are the orthogonal and diagonal matrices, respectively. $\lambda_x$ and $\lambda_x$ are eigenvalues in $x$- and $y$-axes, respectively. The eigenvalues of the calculated covariance is proportional to the variance of the walking motion.


When the factor of the leg kinematic constraint is added, the information matrix in the graph-based Simultaneous Localization And Mapping (SLAM) changes. The negative log likelihood, $\mathbf{F}(\mathbf{x})$ for all observations of graph SLAM is defined as follows \cite{grisetti2010tutorial}:
\begin{equation}
\mathbf{F}(\mathbf{x}) = \sum_{(p,q)}\mathbf{e}_{pq}^T\mathbf{\Omega}_{pq}\mathbf{e}_{pq},
\end{equation}
where $\mathbf{e}_{pq}$ and $\mathbf{\Omega}_{pq}$ are residuals and information matrix between the $p$-th and $q$-th nodes, respectively. This is used as a cost function that is minimized in optimization-based SLAM. At this time, since the leg kinematic residual factor is defined as $\mathbf{\Gamma}$, the cost function is modified as follows:
\begin{equation}
\begin{split}
\mathbf{F^*}(\mathbf{x})
&= \sum_{(p,q)}(\mathbf{\Gamma}\mathbf{e}_{pq})^T\mathbf{\Omega}_{pq}(\mathbf{\Gamma}\mathbf{e}_{pq}) \\
&= \sum_{(p,q)}\mathbf{e}_{pq}^T(\mathbf{\Gamma}^T\mathbf{\Omega}_{pq}\mathbf{\Gamma})\mathbf{e}_{pq} \\
&= \sum_{(p,q)}\mathbf{e}_{pq}^T\mathbf{\Omega}_{pq}^*\mathbf{e}_{pq},
\end{split}
\end{equation}
where $\mathbf{\Omega}_{pq}^* = \mathbf{\Gamma}^T\mathbf{\Omega}_{pq}\mathbf{\Gamma}$, which can be defined as a new information matrix. The residual factor increases as the observations become more accurate. Finally, the adaptive leg kinematic constraint we propose is formulated as follows:
\begin{equation}
\mathbf{r}_{\mathcal{L}_{ij}} = \mathbf{\Gamma}(\mathbf{s}_j - (\mathbf{R}^w_{b_j})^T(\mathbf{R}^w_{b_i}\mathbf{s}_i + \mathbf{t}^w_{b_i} - \mathbf{t}^w_{b_j})).
\end{equation}

\subsection{WALK-VIO for Quadruped Robot}
\begin{figure}[t]
    \centering
    \includegraphics[width=\linewidth]{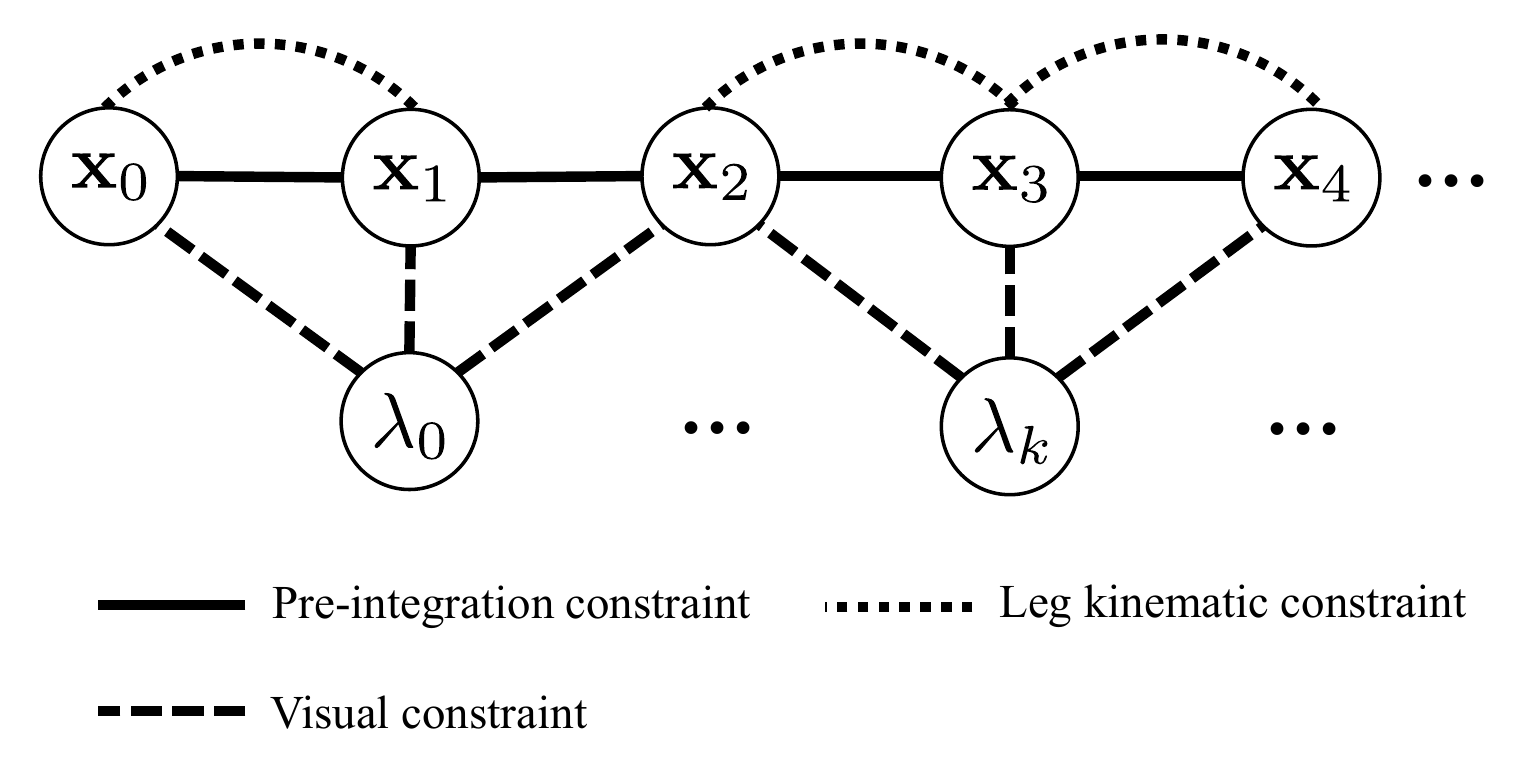}
    \caption{Illustration of the factor graph of \textcolor{black}{WALK-VIO}. The \textcolor{black}{circles indicate} the parameter\textcolor{black}{s} of optimization and \textcolor{black}{are} connected through various constraints.}
    \label{fig:factor_graph}
\end{figure}

The state vector used in our system is as follows:
\begin{equation}
\begin{gathered}
\mathcal{X} = [\mathbf{x}_{0}, \mathbf{x}_{1}, \cdots ,\mathbf{x}_{N-1}, \lambda_{0}, \lambda_{1}, \cdots, \lambda_{n-1}],\\
\mathbf{x}_i = [\mathbf{p}^{w}_{b_i}, \mathbf{q}^{w}_{b_i}, \mathbf{v}^{w}_{b_i}, \mathbf{b}_{a}, \mathbf{b}_{g}],
\label{eq:state}
\end{gathered}
\end{equation}
where $\mathcal{X}$ is the whole state and $\mathbf{x}_i$ is the state in the $i$-th sliding window and is composed of position, quaternion, velocity, and biases of the accelerometer and gyroscope in the appearing order. In addition, the whole state includes the inverse depths of point features represented as $\lambda$. 

Using the state defined in \eqref{eq:state}, the overall cost function for optimization is as follows:
\begin{equation}
\begin{gathered}
\min_{\mathcal{X}} \left\{
\parallel \mathbf{r}_0 \parallel ^2 
+ \sum_{(i,j) \in \mathcal{B}}\parallel{\mathbf{r}_{\mathcal{I}_{ij}}\parallel}_{\mathbf{\Omega}_{\mathcal{I}_{ij}}}^2 
+ \sum_{j \in \mathcal{B}}\sum_{k \in \mathcal{P}}\parallel{\mathbf{r}_{\mathcal{C}_{jk}}\parallel}_{\mathbf{\Omega}_{\mathcal{C}_{jk}}}^2
\right. \\ \left.
+ \sum_{(i,j) \in \mathcal{B}}\parallel{\mathbf{r}_{\mathcal{L}_{ij}}\parallel}_{\mathbf{\Omega}_{\mathcal{L}_{ij}}}^2
\right\},
\label{eq:total residual}
\end{gathered}
\end{equation}
where $\mathbf{r}_0$, $\mathbf{r}_{\mathcal{I}}$, and $\mathbf{r}_{\mathcal{C}}$ represent marginalization, IMU, and camera measurement residual, respectively. In addtion, $\mathbf{\Omega}_{\mathcal{I}}$ and $\mathbf{\Omega}_{\mathcal{C}}$ represent IMU and \textcolor{black}{the} camera measurement information matrices, respectively. These are the same with VINS-Fusion. The newly added terms in the cost function are $\mathbf{r}_{\mathcal{L}}$ and $\mathbf{\Omega}_{\mathcal{L}}$\textcolor{black}{,} which represent leg kinematic measurement residual and the corresponding information matrix. $\mathcal{B}$ and $\mathcal{P}$ mean body states in sliding window and point features, respectively. The factor graph for the defined cost function is shown in Fig.~\ref{fig:factor_graph}.
In the optimization process, Ceres Solver\cite{ceres-solver} was used.

\begin{figure}[t]
    \centering
    \includegraphics[width=\linewidth]{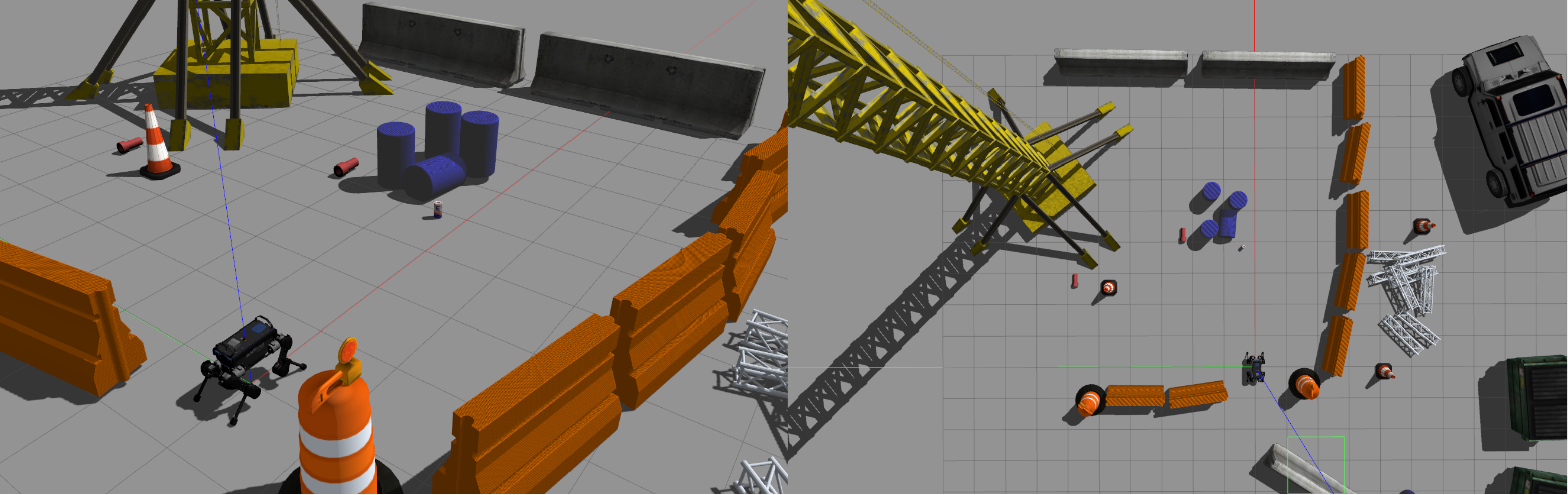}
        \caption{Perspective view (left) and top view (right) of \textcolor{black}{the} simulation environment with ANYmal. The simulation environment is from Champ.}
    \label{fig:simulation_env}
    \vspace{-3mm}
\end{figure}

\begin{figure*}[t]
    \centering
    \subfigure[]{\label{fig:feature_result}\includegraphics[width=0.49\linewidth]{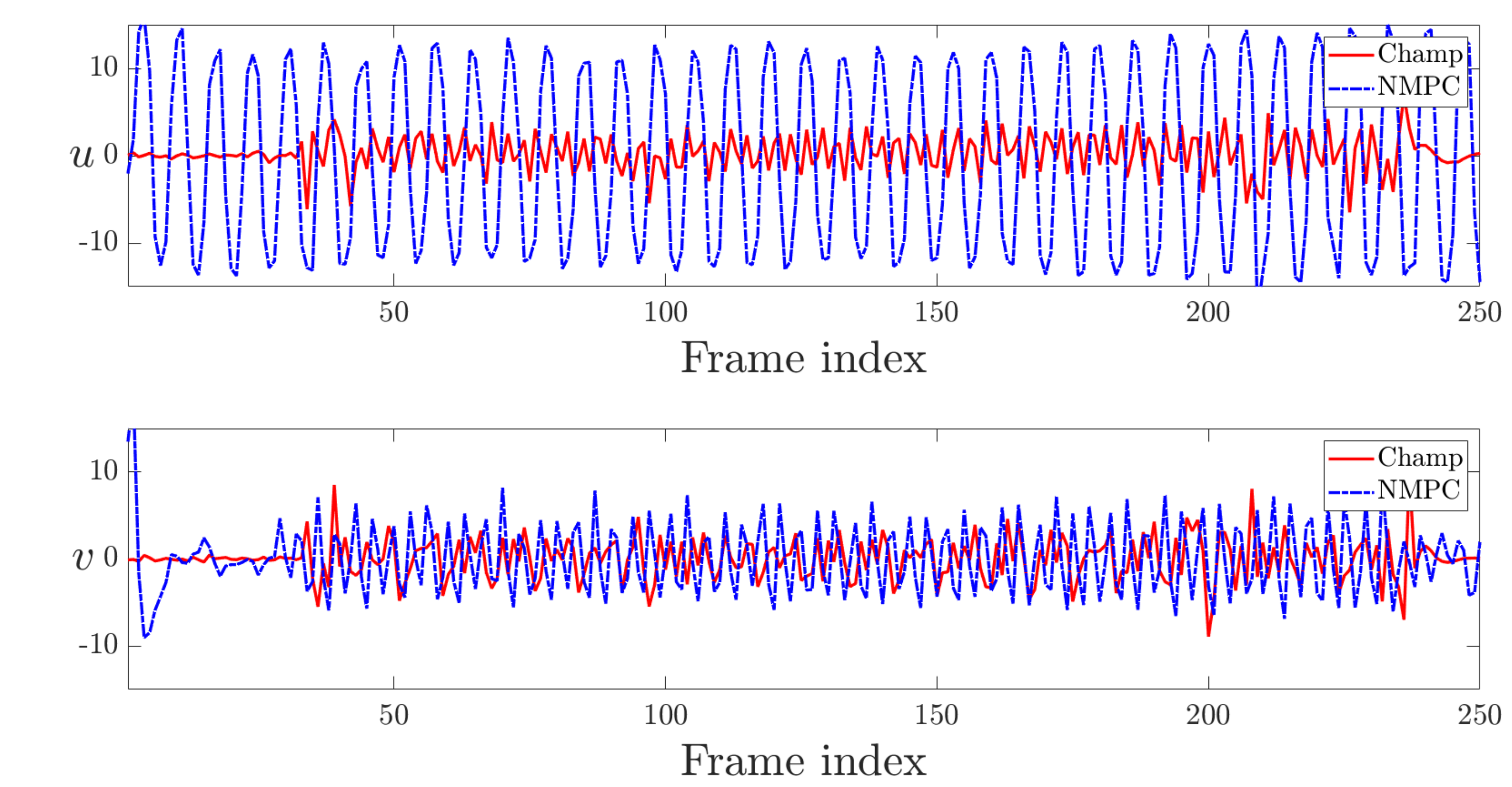}}
    \subfigure[]{\label{fig:eigen_result}\includegraphics[width=0.49\linewidth]{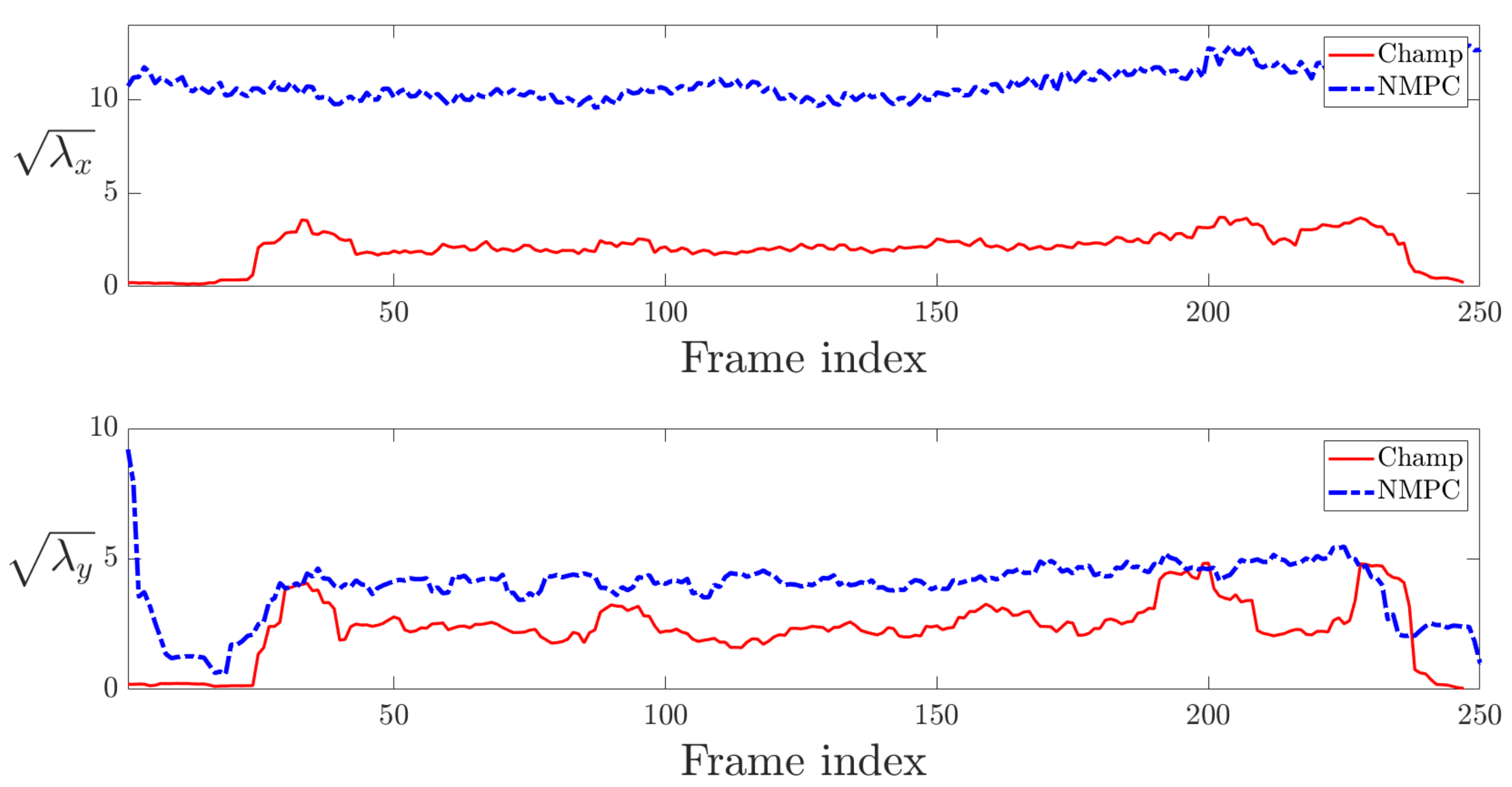}}
    \caption{Plots of \subref{fig:feature_result} average movement of features and \subref{fig:eigen_result} eigenvalues in $x$- and $y$-axes when the quadruped robot goes straight with Champ and NMPC. The calculated factor is multiplied to each leg kinematic measurement residual.}
    \label{fig:feature_graph}
    \vspace{-2mm}
\end{figure*}

\section{Experimental Results}\label{sec:experimental results}
\textcolor{black}{WALK-VIO} is implemented on Ubuntu 18.04 with ROS Melodic. The experiments were conducted with an Intel Core i7-9700K CPU with 32GB of memory.
\subsection{Simulation Setup}
The simulation \textcolor{black}{was} conducted using Gazebo, which can closely interact with ROS. Various quadruped robots can be considered, but in this paper, \textcolor{black}{we used} ANYmal \cite{hutter2016anymal} as the target quadruped robot platform. The stereo camera used Intel RealSense D435i \textcolor{black}{specifications}, and the IMU \textcolor{black}{was} updated at a rate of 100Hz. We used the simulation environment from Champ\cite{champ} as shown in Fig.~\ref{fig:simulation_env}.

In this paper, \textcolor{black}{we used} two controllers to make \textcolor{black}{the} various walking motions of a quadruped robot. First, \textcolor{black}{in the simulated environment}, Champ used three methods to solve the difficulties of locomotion stability, control of ground reaction force, and coordination of four limbs\textcolor{black}{,} based on the method proposed in \cite{lee2013hierarchical}. Leg impedance control provides programmable virtual compliance of each leg, which achieves self-stability in locomotion. The four legs exert forces to the ground using the equilibrium-point hypothesis. A gait pattern modulator imposes the desired footfall sequence. The second controller \textcolor{black}{used was} nonlinear model predictive control (NMPC) \cite{hong2020real}. NMPC \textcolor{black}{treats} a quadruped robot as one rigid body and considers orientation dynamics evolving on the rotation manifold SO(3). Based on that, NMPC is formulated as a constrained nonlinear least-squares problem, which is solved by using an efficient algorithm that enables real-time calculation of optimal solutions. When comparing the two controllers, NMPC yields a relatively larger walking motion. Using these two controllers, we created a dataset \textcolor{black}{for movement} in square, circle, and \textcolor{black}{figure-8} path\textcolor{black}{s}.

\begin{figure*}[t]
    \centering
    \subfigure[]{\label{fig:champ_square}\includegraphics[width=0.32\linewidth]{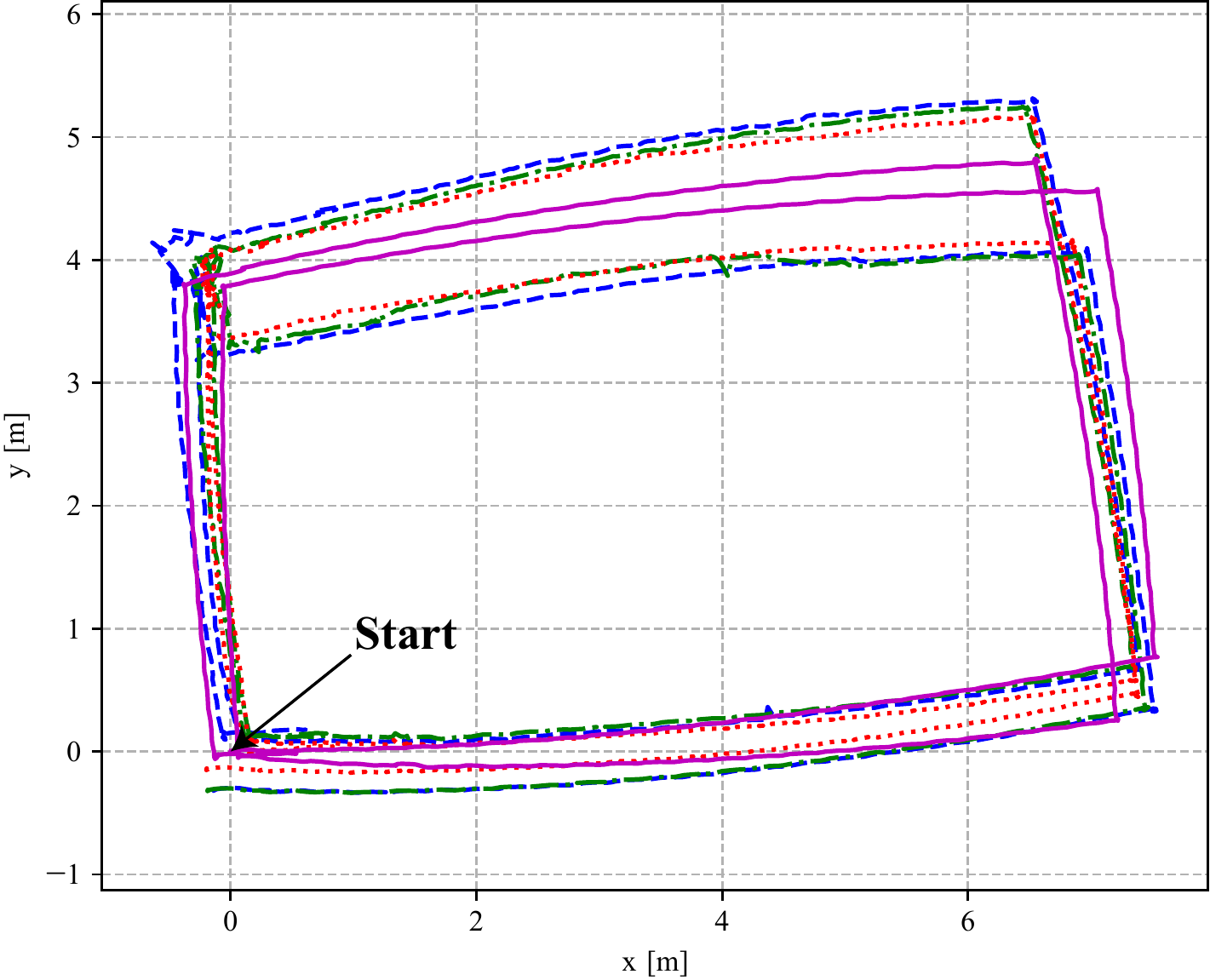}}
    \subfigure[]{\label{fig:champ_circle}\includegraphics[width=0.32\linewidth]{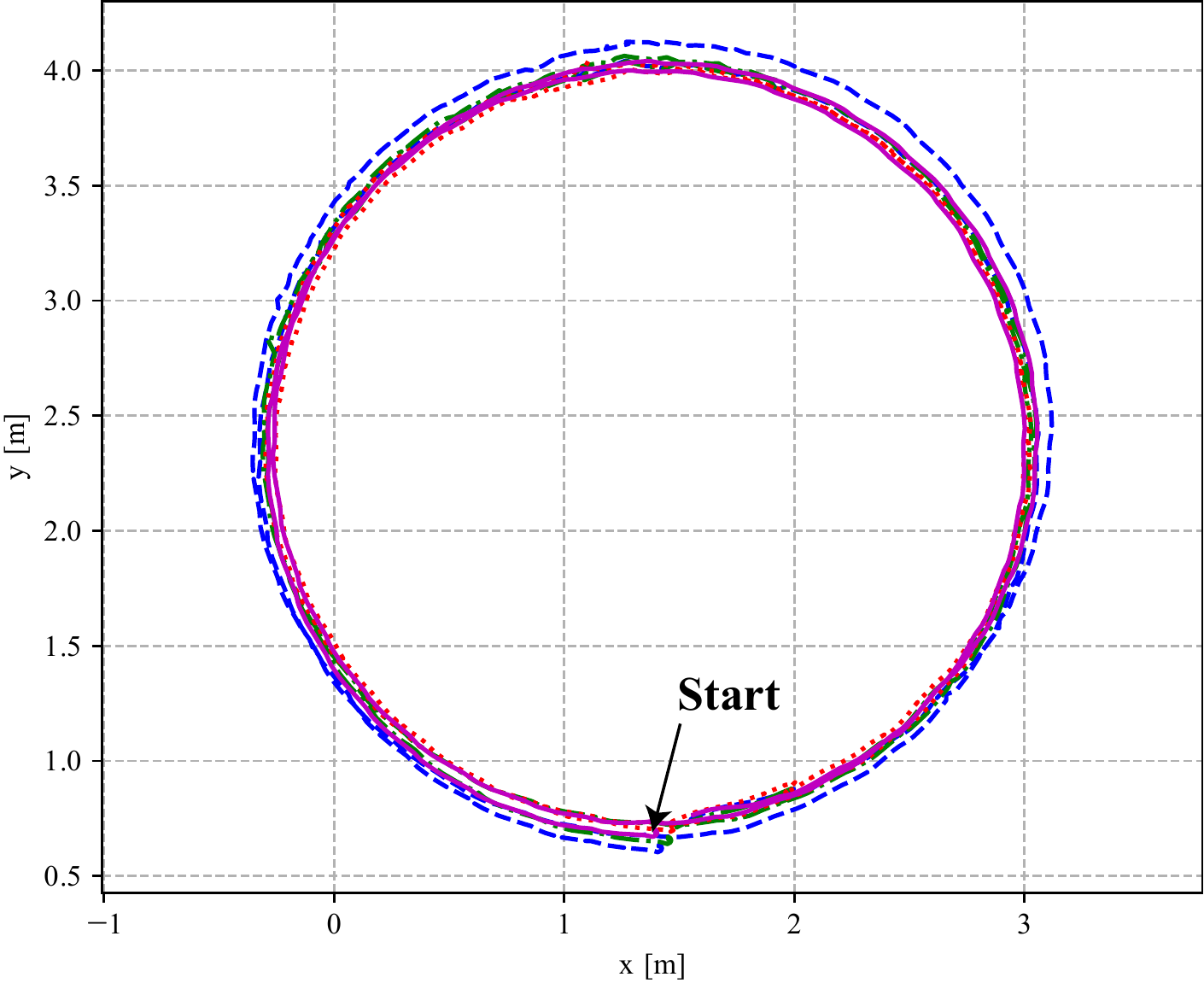}}
    \subfigure[]{\label{fig:champ_eight}\includegraphics[width=0.32\linewidth]{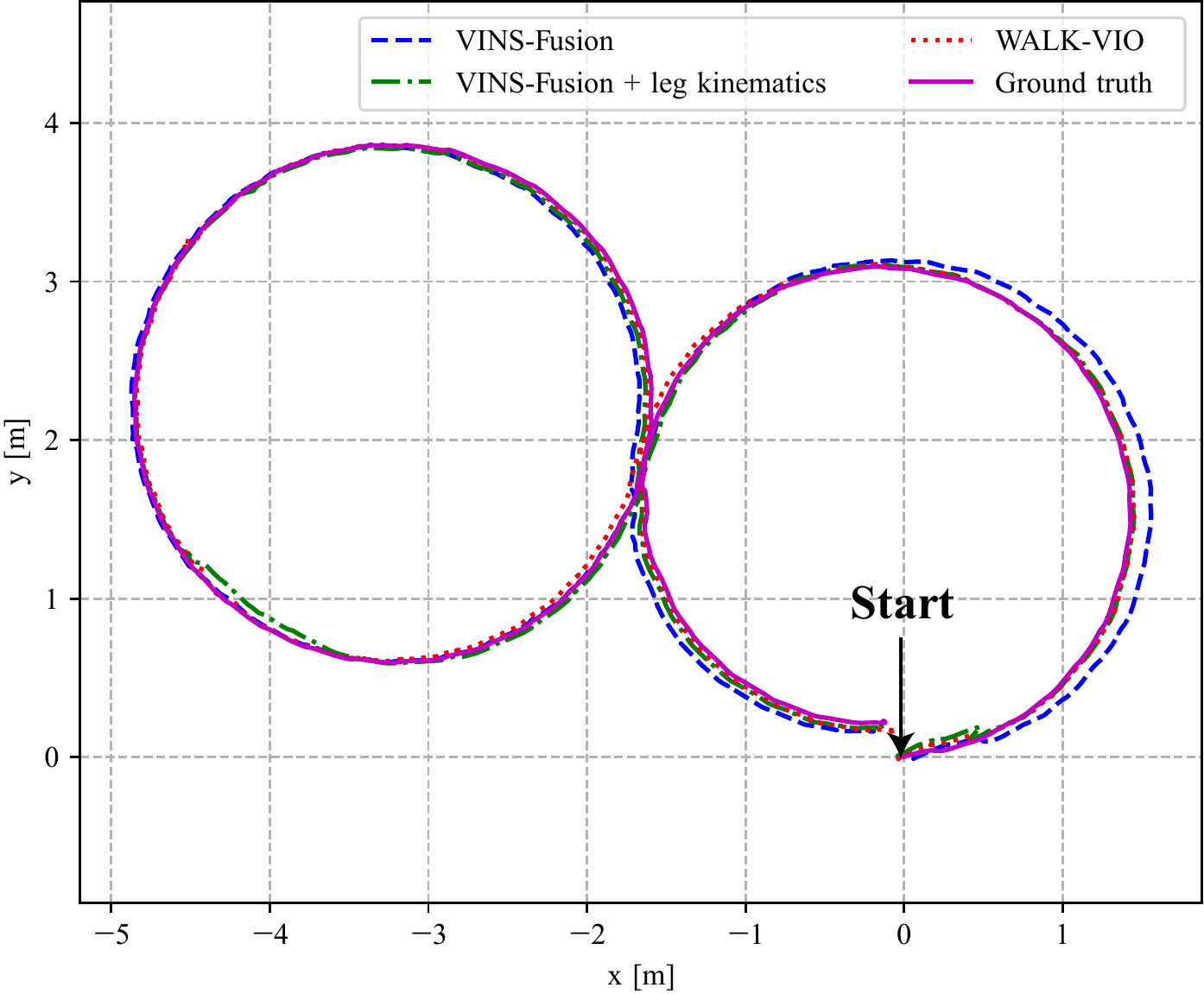}}
    \subfigure[]{\label{fig:hubo_square}\includegraphics[width=0.32\linewidth]{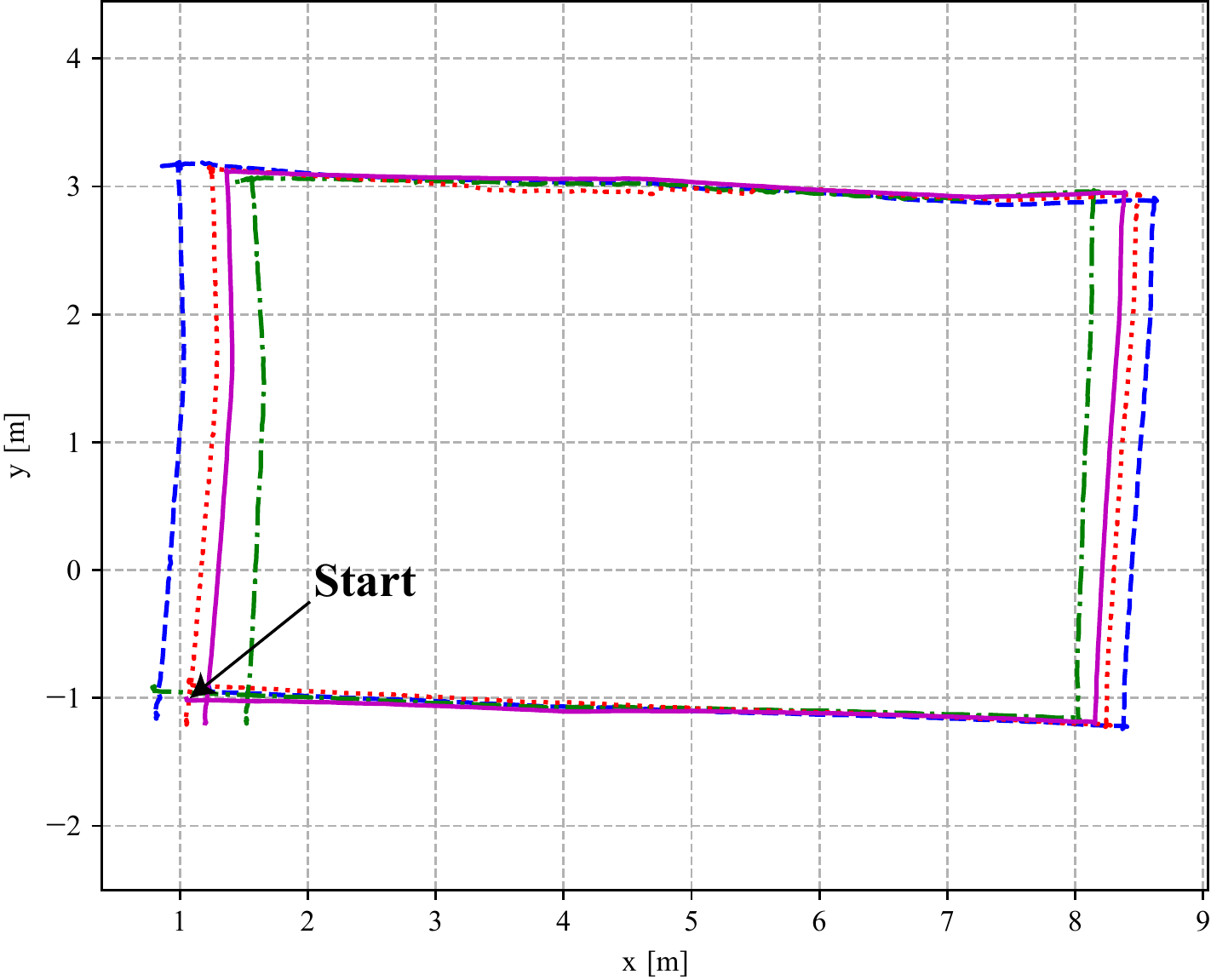}}
    \subfigure[]{\label{fig:hubo_circle}\includegraphics[width=0.32\linewidth]{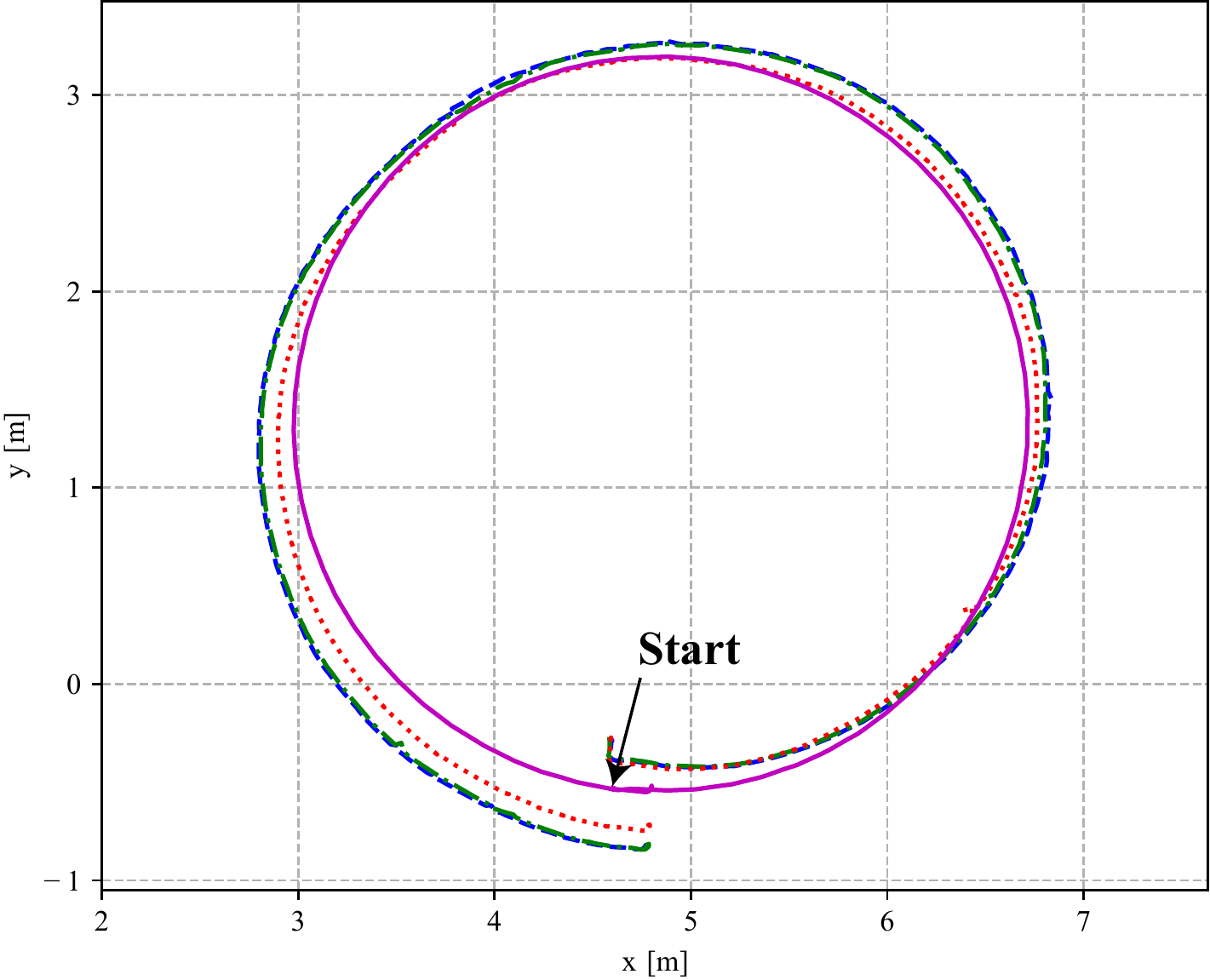}}
    \subfigure[]{\label{fig:hubo_eight}\includegraphics[width=0.32\linewidth]{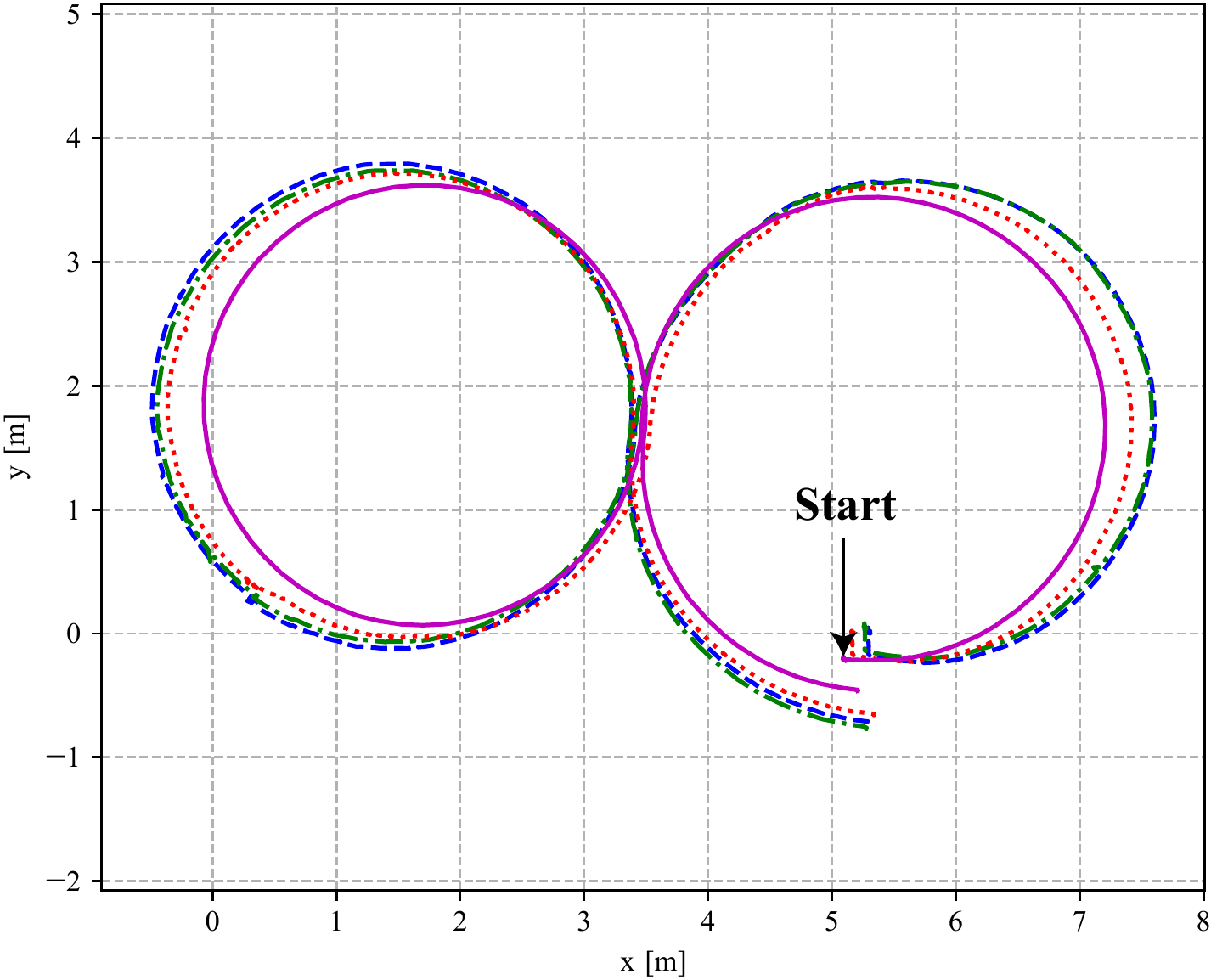}}
    \caption{Top view of trajectories estimated with VINS-Fusion, VINS-Fusion with leg kinematic constraints, and WALK-VIO. \subref{fig:champ_square}~Square path with Champ. \subref{fig:champ_circle}~Circular path with Champ. \subref{fig:champ_eight}~\textcolor{black}{Figure-8} path with Champ. \subref{fig:hubo_square}~Square path with NMPC. 
    \subref{fig:hubo_circle}~Circular path with NMPC. \subref{fig:hubo_eight}~\textcolor{black}{Figure-8} path with NMPC. 
    In general, NMPC has a larger walking motion than Champ.}
    \label{fig:dataset}
    \vspace{-3mm}
\end{figure*}

\begin{figure*}[h]
    \centering
    \includegraphics[width=\linewidth]{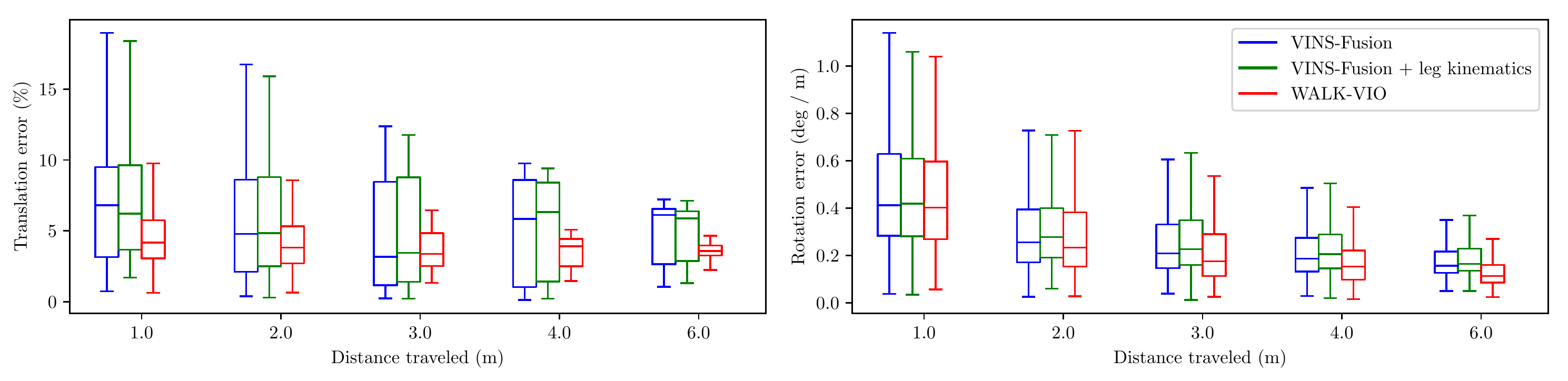}
    \caption{Boxplots of RMSE according to distance traveled for VINS-Fusion, VINS-Fusion with leg kinematic constraints, and WALK-VIO for circular path with NMPC, respectively.}
    \label{fig:boxplot}
\end{figure*}

\subsection{Results}
The average movement of point features and corresponding eigenvalues for ANYmal moving straight with Champ and NMPC are shown in Fig.~\ref{fig:feature_graph}. NMPC, which has a large walking motion, \textcolor{black}{showed} more considerable change in average movement than \textcolor{black}{did} Champ. The eigenvalues in $x$- and $y$-axes are also larger than those of Champ. Finally, the calculated factor was multiplied to the residual, which then was optimized. 

We compare the \textcolor{black}{WALK-VIO} with VINS-Fusion. In addition, \textcolor{black}{we} compared \textcolor{black}{it} with an algorithm that adds a leg kinematic constraint of a fixed factor to VINS-Fusion. For accurate algorithm evaluation, we used the rpg trajectory evaluation tool suggested by Zhang~\textit{et al.} \cite{zhang2018tutorial}. Table~\ref{table:error} shows the translation root mean square error (RMSE) and maximum error for created datasets. In addition, the trajectories for each algorithm are shown in Fig.~\ref{fig:dataset}. In Fig.~\ref{fig:champ_square}, the RMSE of the square path with Champ is larger than other datasets because the robot is close to the wall when rotating. Overall, the algorithms using leg kinematics show better performance than VINS-Fusion. In particular, WALK-VIO shows about 34.5\% smaller error than the VINS-Fusion and about 27.0\% smaller error than the VINS-Fusion with leg kinematics with a fixed factor. Analyzing the dataset with Champ, VINS-Fusion with leg kinematics with a fixed factor and WALK-VIO show 8.5\% and 22.6\% smaller errors than VINS-Fusion, respectively. And for the dataset with NMPC, VINS-Fusion with leg kinematics with a fixed factor and WALK-VIO show 11.7\% and 42.7\% smaller errors than VINS-Fusion, respectively. Through this, it is confirmed that the proposed WALK-VIO is robust even on a large walking motion. Among the results for various datasets, the boxplots for the circular path with NMPC are shown in Fig.~{\ref{fig:boxplot}}. 

\begin{table}[t]
\caption{Translation RMSE and maximum error (Unit: m)}
\label{table:error}
\begin{tabular}{*{10}{c}}\toprule
\multirow{2}{*}{} & \multicolumn{2}{c}{\multirow{2}{*}{VINS-Fusion}} 
                  & \multicolumn{2}{c}{VINS-Fusion} 
                  & \multicolumn{2}{c}{WALK-VIO} \\
                  & \multicolumn{2}{c}{} 
                  & \multicolumn{2}{c}{+ leg kinematics} 
                  & \multicolumn{2}{c}{(Ours)} \\ \midrule
\multirow{1}{*}{Path} & \multirow{2}{*}{RMSE} & \multirow{2}{*}{Max.} & \multirow{2}{*}{RMSE} & \multirow{2}{*}{Max.} & \multirow{2}{*}{RMSE} & \multirow{2}{*}{Max.} \\
\multirow{1}{*}{+ Controller} & & & & & & \\\midrule
\multirow{1}{*}{Square path} & 
\multirow{2}{*}{0.399} & \multirow{2}{*}{0.631} & \multirow{2}{*}{0.386} & \multirow{2}{*}{0.590} & \multirow{2}{*}{\textbf{0.326}} & \multirow{2}{*}{\textbf{0.510}} \\
\multirow{1}{*}{+ Champ} & & & & & &  \\
\multirow{1}{*}{Square path} & 
\multirow{2}{*}{0.306} & \multirow{2}{*}{0.524} & \multirow{2}{*}{0.232} & \multirow{2}{*}{0.359} & \multirow{2}{*}{\textbf{0.124}} & \multirow{2}{*}{\textbf{0.181}} \\
\multirow{1}{*}{+ NMPC} & & & & & &  \\
\multirow{1}{*}{Circular path} & 
\multirow{2}{*}{0.065} & \multirow{2}{*}{0.344} & \multirow{2}{*}{0.061} & \multirow{2}{*}{0.166} & \multirow{2}{*}{\textbf{0.055}} & \multirow{2}{*}{\textbf{0.102}} \\
\multirow{1}{*}{+ Champ} & & & & & &  \\
\multirow{1}{*}{Circular path} & 
\multirow{2}{*}{0.183} & \multirow{2}{*}{0.314} & \multirow{2}{*}{0.180} & \multirow{2}{*}{0.316} & \multirow{2}{*}{\textbf{0.127}} & \multirow{2}{*}{\textbf{0.261}} \\
\multirow{1}{*}{+ NMPC} & & & & & &  \\
\multirow{1}{*}{8-shaped path} & 
\multirow{2}{*}{0.080} & \multirow{2}{*}{0.201} & \multirow{2}{*}{0.051} & \multirow{2}{*}{0.175} & \multirow{2}{*}{\textbf{0.040}} & \multirow{2}{*}{\textbf{0.117}} \\
\multirow{1}{*}{+ Champ} & & & & & &  \\
\multirow{1}{*}{8-shaped path}  & 
\multirow{2}{*}{0.300} & \multirow{2}{*}{0.546} & \multirow{2}{*}{0.285} & \multirow{2}{*}{0.451} & \multirow{2}{*}{\textbf{0.201}} & \multirow{2}{*}{\textbf{0.418}} \\
\multirow{1}{*}{+ NMPC} & & & & & &  \\
\bottomrule 
\end{tabular}
\end{table}

\section{Conclusion}\label{sec:conclusion}
In this paper, \textcolor{black}{we present} a method for state estimation \textcolor{black}{that does not depend} on the controller's performance\textcolor{black}{,} implemented in \textcolor{black}{a} quadruped walking robot. In general, it takes a lot of time and effort to figure out the controller's performance and find the optimal residual factor. We proposed how to find an optimal factor by analyzing the camera movement and how the feature points move, and by appropriately setting each residual factor. In addition, we verified the effectiveness of \textcolor{black}{WALK-VIO} in simulations. However, in order to accurately grasp the effectiveness of \textcolor{black}{WALK-VIO} and obtain realistic results, it \textcolor{black}{must be applied} to an actual quadruped robot. In addition, more diverse experiments on quadruped robots and controllers are needed. 


%




\bibliographystyle{IEEEtran}
\bibliography{reference}
%

%








\end{document}